\documentclass[conference]{IEEEtran}
\IEEEoverridecommandlockouts
\usepackage{cite}
\usepackage{amsmath,amssymb,amsfonts}
\usepackage{graphicx}
\usepackage{textcomp}
\usepackage{xcolor}
\usepackage{booktabs}
\usepackage{url}
\def\BibTeX{{\rm B\kern-.05em{\sc i\kern-.025em b}\kern-.08em
    T\kern-.1667em\lower.7ex\hbox{E}\kern-.125emX}}

\begin{document}
\title{Evaluating and Improving Pedagogical Fit in LLM-Based AI Tutors with the Pedagogical Suitability Index}

\author{\IEEEauthorblockN{Benjamin Barlog, Hudson Craig, Zedong Peng}
\IEEEauthorblockA{\textit{Department of Computer Science} \\
\textit{University of Montana}\\
Missoula, MT, USA \\
\{benjamin.barlog, hudson.craig\}@umconnect.umt.edu, zedong.peng@umt.edu}
}

\maketitle

\begin{abstract}
Large language models (LLMs) are increasingly used as AI tutors, but a correct answer is not always a pedagogically appropriate one. In classroom learning, effective help depends not only on correctness, but also on whether a response matches the learner's current foundation, the course sequence, and the timing of concept introduction. Existing evaluations focus mainly on answer quality, leaving this instructional fit under-measured. We present the Pedagogical Suitability Index (PSI), a composite metric of six theory-informed sub-scores that evaluates how well LLM-generated tutoring responses align with learner readiness and curricular progression, and we further use PSI as a structured feedback signal for response improvement. We evaluate four LLM tutors (ChatGPT, Gemini, Gemma~4, and Qwen~3) across 240 scenario-based evaluations using paired standard and defective prompts, then apply a PSI-guided regeneration protocol to 62 weak-performing cases. Baseline differences across the four tested models were modest overall (PSI range: 0.557 to 0.638), and open-weight and closed models did not exhibit a clear separation in pedagogical fit. Under the tested prompt perturbations, overall PSI remained largely stable ($\Delta = -0.002$), though sub-score trade-offs emerged. More importantly, PSI-guided feedback substantially improved weak-performing cases: 51 of 62 cases improved (82.3\%). Focused manual evaluation of the 62 PSI-selected weak cases provides initial evidence that the identified weaknesses are instructionally meaningful and that many PSI-guided regenerations correspond to human-judged improvement. These results suggest that learner- and curriculum-aware alignment may matter more for effective tutoring than model category alone, and that such alignment is both measurable and improvable.
\end{abstract}

\begin{IEEEkeywords}
AI tutoring, pedagogical fit, LLM evaluation, learner-aligned assessment, metric-guided regeneration, computer science education
\end{IEEEkeywords}

\section{Introduction}

A student asks an AI tutor why a program failed. The tutor responds with a fluent explanation and perhaps even a working fix. Yet the response may still be pedagogically mistimed: it may assume concepts the student has not learned, skip prerequisite reasoning that instruction is intended to build, or provide an answer that resolves the immediate problem without strengthening long-term understanding. In such cases, correctness alone is an incomplete indicator of tutoring quality. For educational use, the central question is not only whether an answer is right, but whether the help is appropriate for this learner at this point in the instructional sequence.

This challenge arises because LLMs typically lack access to crucial instructional context. In many real settings, a model does not know which concepts the learner has already mastered, where the class currently is in the course progression, or how recently related ideas were introduced. As a result, an LLM may produce a generally correct response that is poorly matched to the learner's foundation or to the intended pacing of instruction. This motivates the need for evaluation methods that go beyond factual accuracy and instead measure \textit{pedagogical fit}: whether a response respects prerequisite dependencies, provides suitable scaffolding, aligns with the target cognitive level, and accounts for the timing of concept exposure. Prior work on evaluating AI-generated educational content has focused primarily on factual accuracy or domain-specific quality measures~\cite{yan2024}, without fully capturing these dimensions of learner- and curriculum-aware support. The Zone of Proximal Development~\cite{vygotsky1978} provides theoretical grounding: effective instruction should target the gap between independent capability and guided achievement.

To address this gap, we introduce the Pedagogical Suitability Index (PSI), a theory-informed metric for evaluating whether LLM-generated tutoring responses are appropriate for a learner's current foundation and instructional context. We instantiate this metric on a benchmark of 30 tutoring scenarios, validated by student raters for realism and course fit, and use it to compare four commercial and open-weight LLM tutors under both standard and defective prompting conditions. We then show that PSI can serve not only as an evaluation metric, but also as a structured feedback signal for targeted regeneration, and we further conduct focused manual evaluation of the PSI-selected weak cases to examine whether the identified weaknesses and subsequent improvements are instructionally meaningful.

Our findings suggest that differences in baseline pedagogical fit across current models are present but modest, and do not support a simple open-versus-closed distinction. The more consequential finding is that pedagogical fit can be measured and improved when responses are evaluated against learner readiness, prerequisite structure, and instructional timing. PSI-guided feedback improved 82.3\% of weak cases. All Supplementary materials available at
https://doi.org/10.5281/zenodo.19603300

\section{Related Work}

Prior evaluation work on AI-generated educational responses has focused primarily on answer correctness and text quality~\cite{yan2024, wen2025ai}. AI for personalized learning is well studied~\cite{bodily2017}, and knowledge tracing models~\cite{liu2022, liu2024} can track student mastery over time. In parallel, research on prompt robustness~\cite{zhu2023, dahiya2024leveraging} and self-refinement~\cite{madaan2023, chen2024debug, bai2022} has shown that LLM behavior can shift under noisy inputs and can improve when structured feedback is provided. Student prompts frequently exhibit deficiencies including missing context and incorrect terminology~\cite{denny2024}. These lines of work are valuable, but they do not directly address whether a response is pedagogically appropriate for a learner's current instructional context.

From a learning perspective, pedagogical appropriateness depends on several dimensions that have been studied separately. Work on prerequisite relations~\cite{zuo2020, liang2015, alatrash2025} and knowledge graphs~\cite{gasparetti2018, alzetta2019} highlights the importance of concept ordering. Research on scaffolding~\cite{fernandes2022} and text complexity~\cite{li2025readability} emphasizes the structure and accessibility of instructional support. Bloom's revised taxonomy~\cite{adams2015} focuses on cognitive level, with recent NLP-based automation~\cite{waheed2021}. The Ebbinghaus forgetting curve~\cite{murre2015} points to the importance of timing and retention. Together, these literatures suggest that effective tutoring should be evaluated along multiple learner- and curriculum-aware dimensions, not correctness alone. Related work on similarity measurement~\cite{zhou2022}, activity sequence evaluation~\cite{doroudi2016}, and component interaction analysis~\cite{peng2023resource} further informs our sub-score design.

Our work builds on these strands~\cite{peng2021environment} but differs in two ways. First, it operationalizes these dimensions within a single composite metric for pedagogical fit. Second, it uses that metric not only for diagnosis, but also as a structured signal for targeted response improvement.

\section{The Pedagogical Suitability Index}

PSI is a weighted composite of six normalized sub-scores:
\begin{equation}
\mathrm{PSI} = \sum_{i=1}^{6} w_i \, S_i, \quad S_i \in [0,1], \quad \sum w_i = 1
\end{equation}
Because existing work does not provide a clear empirical basis for weighting these six pedagogical dimensions relative to one another, we adopt equal weights ($w_i = 1/6$) in this study.

\textbf{Knowledge Distance} ($S_{KD}$) measures how well the response targets the learner's zone of proximal development:
$S_{KD} = 1 - | \text{sim}(R, K_{\text{student}}) - \delta_{\text{opt}} |$,
where $\text{sim}(R, K_{\text{student}})$ denotes concept overlap between the response and the student's knowledge state, and $\delta_{\text{opt}} = 0.65$ is the target overlap level, informed by prior work on optimal instructional distance in ZPD-based frameworks. Sensitivity of PSI to this parameter has not been tested in the current study.

\textbf{Temporal Violation} ($S_{TV}$) penalizes responses that introduce concepts before their prerequisites~\cite{doroudi2016}:
$S_{TV} = 1 - \text{violations}/\text{checks}$,
where $\text{violations}$ counts prerequisite-order inversions and $\text{checks}$ denotes the number of prerequisite-order comparisons.

\textbf{Scaffolding Density} ($S_{SD}$) measures the proportion of instructional support elements~\cite{fernandes2022}:
$S_{SD} = (N_{\text{scaffold}}/N_{\text{total}}) \cdot \alpha_{\text{adapt}}$,
where $N_{\text{scaffold}}$ is the number of scaffold indicators detected, $N_{\text{total}}$ is the total number of propositions, and $\alpha_{\text{adapt}} \in [0,1]$ rewards learner-specific adaptation.

\textbf{Ebbinghaus Forgetting} ($S_{EF}$) incorporates retention timing for referenced concepts~\cite{murre2015}:
$S_{EF} = \sum_i \lambda_i \, e^{-t_i / S_i}$,
where $t_i$ is time since last exposure, $S_i$ is memory strength, and $\lambda_i$ is a concept importance weight ($\sum \lambda_i = 1$). In this implementation, $S_{EF}$ depends on scenario metadata rather than response content, yielding identical values across models. We discuss this limitation in Section~VI.

\textbf{Cognitive Load} ($S_{CL}$) penalizes avoidable processing demands~\cite{li2025readability}:
$S_{CL} = 1 - L_{\text{extraneous}}/L_{\text{total}}$,
where $L_{\text{extraneous}}$ denotes extraneous content (filler, hedging, tangential material) and $L_{\text{total}}$ denotes total response length.

\textbf{Bloom's Alignment} ($S_{RA}$) measures cognitive-level match~\cite{adams2015}:
$S_{RA} = 1 - |b_R - b_T|/5$,
where $b_R$ and $b_T$ are the detected and target Bloom levels, respectively.

The current implementation uses proxy measures (regex patterns, keyword matching) rather than deep semantic analysis. We characterize PSI as a theory-informed operationalized benchmark rather than a validated psychometric instrument (see Section~VI).

\section{Experimental Setup}

\subsection{Course Context and Scenario Design}

We designed our evaluation around CSCI~150 (Introduction to Computer Science) at the University of Montana, a 15-week Python-based course for students with no prior programming experience. The curriculum progresses from hardware concepts and variables (Weeks 1--3) through operators, conditionals, and loops (Weeks 4--9) to functions, data structures, NLP, and image processing (Weeks 10--15). This progression from concrete, syntax-focused tasks to abstract, multi-concept integration provides a natural testbed for evaluating AI tutor adaptation across a full semester.

We created 30 structured scenarios (2 per week) spanning Bloom levels 1--4 (Remember: $n$=1, Understand: $n$=7, Apply: $n$=16, Analyze: $n$=6), each with explicit learner profiles, prerequisite maps, and a realistic student question reflecting common misconceptions from actual CSCI~150 tutoring interactions. To validate the scenarios, two undergraduate students who had completed CSCI~150 independently rated each scenario on two 5-point scales: \textit{Realism} (whether the question reflects something a real student would ask) and \textit{Course Fit} (whether the topic and difficulty match the corresponding week). The mean ratings were 3.67 (SD\,=\,0.69) for Realism and 4.10 (SD\,=\,0.60) for Course Fit, with inter-rater agreement within $\pm$1 point at 83.3\% for both criteria. Only one scenario (S14, Mid-term Review -- Debugging) received a mean Realism score below 2.5, suggesting that the scenario set broadly reflects realistic tutoring situations as perceived by students who took the course.

To evaluate prompt robustness, each scenario includes a paired \textit{defective} prompt simulating common student communication patterns. We designed eight defect categories based on observed student behavior: Too Short ($n$=7), Keyword Only ($n$=7), Missing Context ($n$=12), Unclear Goal ($n$=12), Code Dump ($n$=5), Vague Error ($n$=8), Wrong Terminology ($n$=6), and Multi-Issue ($n$=3). These categories capture common ways in which student prompts become instructionally incomplete or noisy, ranging from missing contextual details to ambiguous goals and incorrect technical language. Each scenario may exhibit one or more defect types. This paired design yields $30 \times 2 \times 4 = 240$ total Round~1 evaluations.

\subsection{Concept Taxonomy}

We constructed a prerequisite graph of 85 Python concepts organized hierarchically, from foundational concepts (e.g., \texttt{variables}, \texttt{print\_function}) to advanced topics (e.g., \texttt{dependency\_parsing}, \texttt{tensorflow\_intro}). Each concept has explicitly defined prerequisites; for instance, \texttt{list\_comprehension} requires both \texttt{lists} and \texttt{for\_loop}. This graph encodes the pedagogical structure of the course and enables automated detection of temporal violations in AI responses.

\subsection{AI Tutor Sources}

We evaluated four LLMs spanning commercial and open-weight models:
ChatGPT (OpenAI, GPT-4.5)~\cite{openai2025gpt45},
Gemini (Google, Gemini 3)~\cite{google2026gemini3},
Gemma~4 (Google, 27B, open-weight)~\cite{google2026gemma4}, and
Qwen~3 (Alibaba, 235B-A22B, open-weight)~\cite{qwen2025qwen3}.
Including both commercial and open-weight models allows us to examine whether model category is a meaningful predictor of pedagogical fit. Each system received identical prompts containing student context, prior knowledge, current learning objectives, and the student's question. Responses were collected via standard web interfaces, with one prompt submitted per model at a time to ensure independent responses.

\subsection{PSI Implementation}

The PSI evaluation pipeline operationalizes each sub-score as follows:

\textit{Knowledge Distance} ($S_{KD}$) extracts concepts from each response using keyword matching against the 85-concept taxonomy, computes Jaccard-like overlap with the student's known concepts, and measures deviation from $\delta_{\text{opt}} = 0.65$.

\textit{Temporal Violation} ($S_{TV}$) checks the first-appearance ordering of concepts against the prerequisite graph. For each pair $(c_i, c_j)$ where $c_i$ is prerequisite of $c_j$, we verify that $c_i$ appears before $c_j$ in the response.

\textit{Scaffolding Density} ($S_{SD}$) uses 25 regex-based indicators in five categories: worked examples (code blocks, output demonstrations), step-by-step guidance, hints and prompts, comprehension checks, and analogies. An adaptation multiplier rewards learner-specific references.

\textit{Cognitive Load} ($S_{CL}$) detects extraneous content using patterns for filler phrases (e.g., ``as an AI''), hedging language, and tangential content.

\textit{Bloom's Alignment} ($S_{RA}$) classifies response cognitive level via weighted keyword frequency across six Bloom levels, with bonus weight for executable code examples.

\subsection{PSI-Guided Regeneration Protocol}

We selected 62 weak-performing Round~1 cases for regeneration based primarily on the lowest PSI scores under the defective-prompt condition. We focused on this condition because it represents the more instructionally noisy setting, making it a stronger testbed for targeted pedagogical improvement. The number selected from each model varied because the PSI score distributions differed across models: ChatGPT ($n$=4), Gemini ($n$=23), Gemma~4 ($n$=22), and Qwen~3 ($n$=13). ChatGPT had the fewest weak cases because its baseline PSI distribution was higher overall, leaving fewer cases below the selection threshold. This PSI-based selection targeted the cases with the weakest measured pedagogical fit for Round~2 regeneration. The unequal per-model sample sizes limit strong cross-model conclusions from the regeneration data.

Each regeneration prompt included five components: (a)~the original student context and question, (b)~the model's Round~1 response, (c)~the overall PSI score and all six sub-scores, (d)~a structured diagnosis of specific weaknesses with identified issues and severity, and (e)~explicit improvement requirements formatted as a checklist. For example, a case with low $S_{SD}$ (scaffolding density of 0.05) would receive feedback specifying: ``Add at least one worked code example with expected output, a step-by-step breakdown, and a comprehension check question.'' The regenerated responses were then re-evaluated using the same PSI pipeline.

\subsection{Manual Validation}

We conducted focused manual evaluation on the 62 PSI-selected weak cases used in Round~2 regeneration. For each case, a course instructor with prior experience teaching CSCI~150 examined both the original Round~1 response and the regenerated Round~2 response against the course materials (29 slide decks, 31 Colab notebooks). This evaluation assessed two criteria: \textit{Meaningful Improvement} (whether the regenerated response demonstrates better scaffolding structure than the original) and \textit{Weaknesses Resolved} (whether the specific PSI-identified issues have been adequately addressed). Because this manual evaluation was limited to the PSI-selected weak cases and was performed by a single expert rater, the results should be interpreted as preliminary.

\section{Results}

\subsection{Cross-Model Comparison (Round~1)}

\begin{table}[htbp]
\caption{Mean PSI Sub-Scores (Standard Prompts, $n$=30 per model)}
\label{tab:cross}
\centering
\begin{tabular}{lcccccc|c}
\toprule
\textbf{Model} & $S_{KD}$ & $S_{TV}$ & $S_{SD}$ & $S_{EF}$ & $S_{CL}$ & $S_{RA}$ & \textbf{PSI} \\
\midrule
ChatGPT & .662 & .578 & .492 & .334 & .996 & .767 & \textbf{.638} \\
Qwen~3  & .618 & .606 & .271 & .334 & .999 & .740 & .595 \\
Gemma~4 & .627 & .502 & .185 & .334 & 1.00 & .740 & .565 \\
Gemini  & .632 & .535 & .087 & .334 & 1.00 & .753 & .557 \\
\bottomrule
\end{tabular}
\end{table}

As shown in Table~\ref{tab:cross}, PSI scores ranged from 0.557 (Gemini) to 0.638 (ChatGPT), with Qwen~3 (0.595) and Gemma~4 (0.565) in between. The overall spread of 0.081 points is modest, and model type (commercial versus open-weight) did not cleanly predict PSI rank: the open-weight Qwen~3 outperformed the commercial Gemini. The primary differentiator across models is $S_{SD}$: ChatGPT (0.492) provides over five times the scaffolding of Gemini (0.087), suggesting that scaffolding behavior varies substantially even when overall pedagogical fit is broadly comparable. Two sub-scores show limited discriminative power in the current implementation: $S_{EF}$ is identical across models (0.334) because it depends on scenario metadata, and $S_{CL}$ is near-saturated ($\geq$0.996).

\begin{figure}[htbp]
\centering
\includegraphics[width=0.85\columnwidth]{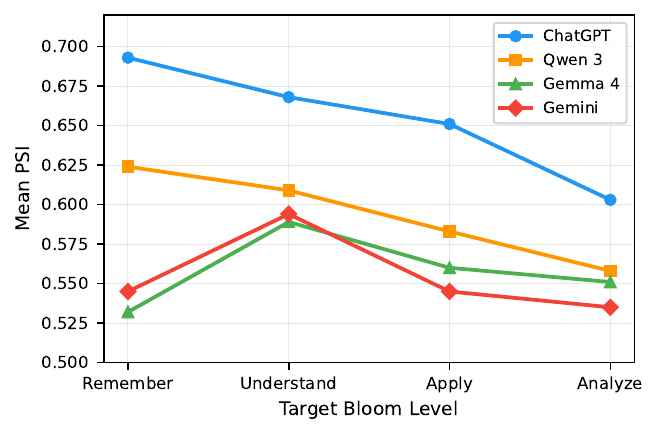}
\caption{Mean PSI by target Bloom level across all four models. All models show a declining trend as cognitive complexity increases.}
\label{fig:bloom}
\end{figure}

Fig.~\ref{fig:bloom} shows that PSI decreases with target Bloom level across all models, from a cross-model mean of 0.60 at Remember to 0.56 at Analyze. This trend is consistent with the pedagogical fit perspective: as prerequisite chains lengthen and target cognitive levels rise, maintaining learner-appropriate responses becomes harder for all models.

\subsection{Prompt Robustness}

Table~\ref{tab:robust} compares PSI under standard versus defective prompts.

\begin{table}[htbp]
\caption{PSI Under Standard vs.\ Defective Prompts ($n$=30 per cell)}
\label{tab:robust}
\centering
\begin{tabular}{lccc}
\toprule
\textbf{Model} & \textbf{Standard} & \textbf{Defective} & $\boldsymbol{\Delta}$ \\
\midrule
ChatGPT & 0.638 & 0.656 & $+$0.018 \\
Qwen~3  & 0.595 & 0.576 & $-$0.018 \\
Gemma~4 & 0.565 & 0.563 & $-$0.002 \\
Gemini  & 0.557 & 0.552 & $-$0.005 \\
\midrule
\textbf{Overall} & 0.589 & 0.587 & $-$0.002 \\
\bottomrule
\end{tabular}
\end{table}

Across the tested scenarios, overall PSI changed only marginally ($\Delta = -0.002$), although sub-score trade-offs emerged (Table~\ref{tab:robust_sub}).

\begin{table}[htbp]
\caption{Sub-Score Impact of Defective Prompts (All Models, $n$=120)}
\label{tab:robust_sub}
\centering
\begin{tabular}{lccc}
\toprule
\textbf{Sub-Score} & \textbf{Standard} & \textbf{Defective} & $\boldsymbol{\Delta}$ \\
\midrule
$S_{KD}$ & 0.634 & 0.593 & $-$0.041 \\
$S_{TV}$ & 0.555 & 0.615 & $+$0.059 \\
$S_{SD}$ & 0.259 & 0.216 & $-$0.043 \\
$S_{EF}$ & 0.334 & 0.334 & $\phantom{+}$0.000 \\
$S_{CL}$ & 0.999 & 0.999 & $\phantom{+}$0.000 \\
$S_{RA}$ & 0.750 & 0.763 & $+$0.013 \\
\bottomrule
\end{tabular}
\end{table}

The sub-score trade-offs are consistent with the pedagogical fit perspective: $S_{TV}$ improved ($+0.059$), suggesting models compensate for missing context by imposing stronger prerequisite ordering, while $S_{KD}$ ($-0.041$) and $S_{SD}$ ($-0.043$) both decreased, reflecting difficulty calibrating learner-specific knowledge distance and scaffolding without explicit context.

\subsection{PSI-Guided Regeneration (Round~2)}

Table~\ref{tab:regen} summarizes regeneration results.

\begin{table}[htbp]
\caption{Round~2 Regeneration Results}
\label{tab:regen}
\centering
\begin{tabular}{lcccccc}
\toprule
\textbf{Model} & $n$ & \textbf{R1} & \textbf{R2} & $\boldsymbol{\Delta}$ & \textbf{Impr.} & \textbf{Decl.} \\
\midrule
ChatGPT & 4  & .543 & .629 & $+$.086 & 4/4   & 0/4 \\
Gemini  & 23 & .533 & .599 & $+$.066 & 22/23 & 1/23 \\
Gemma~4 & 22 & .541 & .580 & $+$.039 & 16/22 & 6/22 \\
Qwen~3  & 13 & .534 & .562 & $+$.028 & 9/13  & 4/13 \\
\midrule
\textbf{All} & 62 & .537 & .586 & $+$.049 & 51/62 & 11/62 \\
\bottomrule
\end{tabular}
\end{table}

PSI-guided feedback produced improvement in 51 of 62 cases (82.3\%), with a mean gain of $+0.049$ PSI points. All four models improved on average, though unequal sample sizes (4 to 23 per model) limit strong cross-model conclusions.

Table~\ref{tab:regen_sub} shows the sub-score changes.

\begin{table}[htbp]
\caption{Sub-Score Changes in Regeneration (Mean of 62 Cases)}
\label{tab:regen_sub}
\centering
\begin{tabular}{lccc}
\toprule
\textbf{Sub-Score} & \textbf{R1} & \textbf{R2} & $\boldsymbol{\Delta}$ \\
\midrule
$S_{KD}$ & 0.593 & 0.632 & $+$0.039 \\
$S_{TV}$ & 0.586 & 0.553 & $-$0.034 \\
$S_{SD}$ & 0.073 & 0.345 & $+$0.272 \\
$S_{EF}$ & 0.291 & 0.291 & $\phantom{+}$0.000 \\
$S_{CL}$ & 1.000 & 1.000 & $\phantom{+}$0.000 \\
$S_{RA}$ & 0.752 & 0.697 & $-$0.055 \\
\bottomrule
\end{tabular}
\end{table}

The dominant driver is $S_{SD}$, which increased from 0.073 to 0.345 ($+0.272$), the largest single sub-score change in the study. This suggests that scaffolding is the dimension most responsive to structured feedback: when the regeneration prompt explicitly requests worked examples, step-by-step breakdowns, and comprehension checks, models incorporate them. $S_{KD}$ also improved ($+0.039$), suggesting better knowledge calibration in regenerated responses. However, $S_{TV}$ declined ($-0.034$) and $S_{RA}$ decreased ($-0.055$), indicating that adding scaffolding content can sometimes disrupt prerequisite ordering or shift the response's cognitive level. $S_{EF}$ remained unchanged, consistent with its metadata-dependent nature. These patterns suggest that single-round feedback is most effective for concrete, instruction-like dimensions (scaffolding), with diminishing impact on dimensions requiring deeper pedagogical judgment. This trade-off also suggests that the observed improvements reflect targeted responses to specific feedback instructions rather than uniform optimization of the composite score, and that multi-round feedback balancing multiple dimensions may be needed to avoid such trade-offs.

\subsection{Manual Validation}

Table~\ref{tab:manual} presents the focused manual evaluation results for the 62 PSI-selected weak cases.

\begin{table}[htbp]
\caption{Manual Evaluation of 62 PSI-Selected Weak Cases}
\label{tab:manual}
\centering
\begin{tabular}{lcc}
\toprule
\textbf{Criterion} & \textbf{Score} & \textbf{Rate} \\
\midrule
Meaningful Improvement & 46/62  & 74.2\% \\
Weaknesses Resolved    & 51/62  & 82.3\% \\
\bottomrule
\end{tabular}
\end{table}

The ``weaknesses resolved'' rate (82.3\%) matched the automated PSI improvement count (51/62), providing initial evidence that PSI-indicated improvement often corresponds to human-judged pedagogical gains. The ``meaningful improvement'' rate was lower (74.2\%), indicating that some PSI gains were not clearly visible instructionally.

\section{Discussion}

\subsection{PSI as a Measure of Pedagogical Fit}

Our results suggest that PSI captures meaningful differences in how well LLMs align their responses with learner readiness and curricular context. As a diagnostic tool, it identifies scaffolding density ($S_{SD}$) as the primary differentiator, with a 5.7$\times$ variation between the highest (ChatGPT, 0.492) and lowest (Gemini, 0.087) performers. As a feedback signal, structured PSI diagnostics improve 82.3\% of weak cases, with $S_{SD}$ showing the largest gain ($+0.272$). This dual use, measuring pedagogical fit and then translating that measurement into targeted improvement, is the central contribution.

\textit{Methodological note.} PSI should be interpreted as an operationalized, theory-informed benchmark rather than a definitive psychometric instrument. Two sub-scores ($S_{EF}$, $S_{CL}$) show limited discriminative power in the current implementation: $S_{EF}$ is identical across models because it depends on scenario metadata, and $S_{CL}$ is near-saturated ($\geq$0.996). In the current implementation, most cross-model differentiation comes from the remaining four sub-scores. PSI's value lies in enabling consistent, curriculum-aware comparison across models and generating structured feedback for improvement, while future work should refine component validity, weighting schemes, and inter-rater reliability.

\subsection{Comparison with Prior Work}

PSI contributes to an emerging line of work on composite evaluation of AI-generated tutoring content, simultaneously considering prerequisite ordering, scaffolding quality, retention timing, cognitive load, and Bloom alignment. Prior frameworks have focused on isolated dimensions: factual accuracy~\cite{yan2024}, text readability~\cite{li2025readability}, or knowledge tracing~\cite{liu2022}. PSI unifies these under a single interpretable score while maintaining diagnostic capability through sub-score analysis. The regeneration protocol extends recent work on self-refinement~\cite{madaan2023} by showing that domain-specific pedagogical feedback can produce targeted, measurable improvements.

\subsection{Prompt Robustness}

Under the tested prompt perturbations, overall PSI remained largely stable, although sub-score trade-offs emerged, particularly between prerequisite ordering ($S_{TV}$: $+0.059$) and knowledge calibration ($S_{KD}$: $-0.041$). Models appear to compensate for missing context by imposing default pedagogical structure, maintaining instructional ordering at the cost of learner-specific calibration. These findings are limited to the 30 tested scenarios and eight defect categories; broader generalization requires further study.

\subsection{Implications for AI System Design}

Three implications follow from these results. First, scaffolding is the most actionable improvement target: it varies substantially across models (0.087 to 0.492) and responds strongly to structured feedback ($S_{SD}$: $+0.272$). Second, retention-aware support likely requires external learner-history data, since $S_{EF}$ remained unchanged across both models and rounds. Third, the fact that all four models improved on average under PSI-guided feedback suggests that curriculum-aware alignment may matter more for tutoring quality than model category alone.

\subsection{Limitations}

Four categories of limitation should be noted. \textit{Measurement}: PSI relies on proxy measures (regex, keyword matching) rather than deep semantic analysis, and two sub-scores ($S_{EF}$, $S_{CL}$) show limited discriminative power in the current implementation. \textit{Scope}: the study is based on one course (CSCI~150), one concept taxonomy, and 30 scenarios; generalization to other courses and disciplines remains to be established. \textit{Experimental design}: results reflect a single snapshot of each model under specific interface and parameter settings, the regeneration protocol uses a single feedback round with unequal per-model sample sizes (4 to 23), and we report descriptive comparisons without inferential statistics. \textit{Validation}: focused manual evaluation covered only the 62 PSI-selected weak cases and was performed by a single expert rater without inter-rater reliability measurement.

\section{Conclusion}

We presented PSI, a six-component metric for evaluating and improving the pedagogical fit of LLM-based AI tutors. Across 240 automated evaluations and 62 regeneration cases, pedagogical fit differed across models, but not in a way that supports a simple ranking narrative or a clear open-versus-closed divide. Instead, the main finding is that effective AI tutoring depends less on model category alone and more on whether responses are aligned with learner readiness, prerequisite structure, cognitive level, and instructional timing. Within this study, baseline PSI ranged from 0.557 to 0.638, overall performance remained largely stable under defective prompts, and PSI-guided feedback improved 82.3\% of weak cases, with scaffolding density showing the largest gain ($+0.272$). Focused manual evaluation of the PSI-selected weak cases provides initial evidence that PSI-based improvement corresponds to human-judged pedagogical gains.

PSI should be understood as a theory-informed operationalized benchmark with known limitations, including equal weighting, proxy-based sub-score computation, and two components ($S_{EF}$, $S_{CL}$) with limited discriminative power in the current implementation. Its value lies in enabling consistent cross-model comparison on pedagogical dimensions and, more importantly, in translating diagnostic measurement into structured feedback that demonstrably improves learner-aligned output quality. These results suggest that pedagogical fit is both measurable and improvable through curriculum-aware evaluation. Future work should test PSI in live tutoring settings with authentic student-AI interactions and examine whether the same patterns hold across additional courses and measured learning outcomes. These extensions would help establish whether the pedagogical-fit perspective generalizes beyond the present benchmark.
\bibliographystyle{IEEEtran}
\bibliography{references}

\end{document}